\documentclass[10pt,twocolumn]{article}

\usepackage[T1]{fontenc}
\usepackage[utf8]{inputenc}
\usepackage{amsmath,amssymb}   % must precede newtxmath (avoids \Bbbk clash)
\usepackage{newtxtext}
\usepackage{newtxmath}
\usepackage[a4paper,top=2.4cm,bottom=2.2cm,left=1.9cm,right=1.9cm,columnsep=0.62cm]{geometry}
\usepackage{microtype}
\usepackage{balance}

\usepackage{booktabs}
\usepackage{tabularx}
\usepackage{array}
\usepackage{graphicx}
\usepackage{caption}
\usepackage{subcaption}
\usepackage{xcolor}
\usepackage{tikz}
\usetikzlibrary{arrows.meta,positioning,shapes.geometric,fit,backgrounds,calc}
\usepackage{enumitem}
\usepackage{etoolbox}
\usepackage[numbers,sort&compress]{natbib}
\usepackage{fancyhdr}
\usepackage{xurl}
\usepackage[colorlinks=true,allcolors=blue!55!black,bookmarksnumbered=true]{hyperref}

\definecolor{gblue}{RGB}{31,78,121}
\definecolor{ggold}{RGB}{176,132,0}
\definecolor{gwarn}{RGB}{150,58,42}
\definecolor{grey}{RGB}{245,246,248}

\newcommand{\qzero}{Q0}
\newcommand{\qhard}{\ensuremath{\mathrm{Q0\text{-}hard}}}
\newcommand{\pp}{\ensuremath{\,\mathrm{pp}}}

\title{\vspace{-1.1cm}\textbf{Beyond Linear Context:\\ Graph-Guided Evidence Navigation for Long-Novel Reasoning\\ with a Local 9B Language Model}}

\author{%
  \normalsize\textbf{Wenji Fu}\\[3pt]
  \footnotesize Research Institute of Economics and Management\\
  \footnotesize Southwestern University of Finance and Economics\\
  \footnotesize\texttt{fuwenji61616@gmail.com}
}
\date{}

\begin{document}
\maketitle
\thispagestyle{fancy}

\begin{abstract}
\noindent
Long-context models read a novel the way a person reads a printout: one token after
another, in narrative order, with the whole history competing for a fixed budget of
attention. A detective does not work that way. They sort what happened when, and they
keep a map of who relates to whom, so a clue from chapter one can meet a question asked
at the end of the book. We test whether a frozen knowledge graph can give a small local
model that same freedom. Thirty detective novels and 234 multiple-choice questions are
answered by one fixed \texttt{qwen3.5:9b} reader under nine conditions: five graph
routes, a recent-window baseline, whole-book compression, ordinary vector retrieval, and
a question-only control. The strongest graph route reaches 53.85\% (126/234) against
46.15\% for the recent window, 51.28\% for compression, 51.71\% for vector retrieval and
40.17\% for question-only. On the subset that no model can answer without the book, the
graph route reaches 42.86\%. None of the fifteen graph--baseline contrasts survives
Holm correction, so we present the result as exploratory evidence about a design, not as
a confirmed gain. Two structural findings survive scrutiny better than the headline
number: annotated evidence concentrates in the topological core of these graphs
($2.35\times$ enrichment, pooled), and the two graph-building pipelines differ so much in
annotation coverage (16\% versus 73\% of clue paragraphs) that pooled accuracy alone
would hide which bottleneck is being measured.
\end{abstract}

\vspace{-2pt}
\noindent\textbf{Keywords:} long-context reasoning \textperiodcentered\ knowledge graphs
\textperiodcentered\ retrieval-augmented generation \textperiodcentered\ small language
models \textperiodcentered\ narrative question answering \textperiodcentered\
pre-registered evaluation

\section{Introduction}
\label{sec:intro}

Ask a reader about a novel they finished last month and they will not replay it from page one. They will sort what happened when, remember who was
connected to whom, and jump straight to the passages that matter. The evidence is the
same either way; the difference is the order in which it is reached.

A transformer reading the same novel has no such freedom. Tokens enter the context as a
linear history, and what the model can use at step $t$ depends on the position of the
evidence as much as on its relevance \citep{liu2024lost}. Detective fiction makes the problem vivid. The clue that settles the case often sits a hundred pages away from the question, under a different name for the same person, in a paragraph that looks like
scenery. DetectiveQA was built around exactly this difficulty, with novel-length
contexts and questions that require several pieces of evidence
\citep{xu2024detectiveqa}.

This paper asks a narrow version of a large question. If the evidence is reorganized
into a graph that a retrieval policy can walk, does a small local model answer
long-novel questions better than it does from a linear context of the same budget? We
freeze thirty novel graphs, hold the reader fixed at \texttt{qwen3.5:9b} with reasoning
disabled, and run nine conditions on every question. Five are graph routes, three are
baselines, and a question-only control. Gold clue and answer paragraphs are withheld
from every method and used only to audit what was retrieved.

We report four things.

\begin{itemize}[leftmargin=1.05em,itemsep=1.5pt,topsep=2pt]
\item Accuracy for all nine conditions on 234 questions, with Wilson intervals, and the
same values split by the two graph-building cohorts (Section~\ref{sec:results}).
\item Paired comparisons against each baseline on the same questions, with
novel-clustered bootstrap intervals and exact McNemar tests, corrected across the
planned family.
\item An evidence audit that separates two questions that pooled accuracy blends
whether annotated evidence sits in the structurally central part of a graph, and
whether the graph-building pipeline maps the annotation into the graph at all.
\item A difficulty census that shows where the graph routes are the only conditions to
answer correctly, and how many items nobody answers.
\end{itemize}

The headline number is 53.85\% for the strongest graph route against 46.15\% for a
recent-window baseline, 51.28\% for whole-book compression and 51.71\% for ordinary
vector retrieval. The paired contrast against the recent window is $+7.69\pp$
($p=0.0505$ before correction, $0.76$ after Holm correction across fifteen planned
contrasts), so we treat it as a design signal rather than a demonstrated effect. The
two structural findings hold up better. Annotated evidence is $2.35\times$ enriched in
the graph 2-core, and the two cohorts map clue paragraphs into their graphs at 16\% and
73\% respectively. Any accuracy comparison that pools those cohorts measures the
builders as much as the methods.

\section{Related Work}
\label{sec:related}

\paragraph{How models use long contexts.}
Position matters. Retrieval quality in the middle of a long input is measurably worse than at the ends \citep{liu2024lost}, and the effect survives changes in model size.
Work on attention calibration and position-agnostic training attacks the problem inside
the model \citep{hsieh2024found}; we leave the model alone and change what it is
allowed to read.

\paragraph{Compression, retrieval, and graph structure.}
Standard retrieval-augmented generation hands the reader a handful of passages
\citep{lewis2020rag}, and LongRAG scales the retrieved unit rather than the retriever
\citep{zhao2024longrag}. RAPTOR builds a tree of recursive summaries
\citep{sarthi2024raptor}. Graph-based approaches add structure to the index. HippoRAG
runs personalized PageRank over a knowledge graph for multi-hop integration
\citep{gutierrez2024hipporag}, GraphRAG summarizes graph communities for query-focused
summarization \citep{edge2024graphrag}, and GraphReader lets an agent plan its own
exploration of a graph under a small context window \citep{li2024graphreader}. Chain-of
thought prompting is the single-model analogue of stepwise evidence gathering
\citep{wei2022chain}. Our setting is deliberately narrower than any of these.
one local 9B reader, graphs frozen before scoring, multiple-choice questions over full
novels, and no graph rebuilding during evaluation.

\paragraph{What this paper adds.}
Most of the graph-navigation literature reports accuracy. We add an audit of the index
itself, because a graph can be structurally rich and still fail to contain the evidence
a question needs. That distinction, and the cohort split that exposes it, is the part of
this study we would keep if the accuracy comparison were inconclusive.

\section{Data and Provenance}
\label{sec:data}

\paragraph{Corpus.}
The evaluation uses DetectiveQA, thirty novels in Chinese translation with 234
multiple-choice questions with four options each, and paragraph-level gold annotations
for the clue and the final answer \citep{xu2024detectiveqa}. Each question keeps its
official gold option and its annotated clue and answer paragraphs.

\paragraph{Frozen graphs, two builders.}
All thirty graphs already existed when this evaluation began, and none was rebuilt. The
first cohort, \emph{old20}, comes from a legacy pipeline built with a
\texttt{qwen2.5:7b} extractor; the second, \emph{new10}, comes from a Pass2-v4
relation-centred pipeline built with \texttt{qwen3.5:9b}. Graph size ranges from 257 to
1,071 nodes (median 427) and from 210 to 1,078 edges (median 560). The frozen manifest
records, for every novel, the absolute path, builder lineage, node and edge counts, file
size, modification time and SHA-256 hash; hashes are checked before and after each new
run. Incomplete replacement graphs for novels 26--28 are excluded from every number
reported here.

The two cohorts differ in ways that matter for interpretation, so we report them separately
throughout. Pooled values appear in the tables, labelled descriptive.

\paragraph{Reader and decoding.}
Every newly executed condition answers with the same local \texttt{qwen3.5:9b} model
through Ollama, reasoning disabled, fixed prompts, and a common context budget. No
external API is called. Each answer record keeps the chosen option, parse status, the
evidence payload, estimated input tokens where exact legacy accounting exists, elapsed
time for new calls, and a run signature. Failed items stay in the denominator.

\begin{figure*}[t]
\centering
\resizebox{\textwidth}{!}{% Provenance chain --- TikZ body only. The float environment and caption live in the
% calling section file (03_data.tex), so this file must NOT open a figure environment.
\begin{tikzpicture}[
  node distance=4mm,
  box/.style={draw=black!35,rounded corners=1.6pt,fill=blue!4,align=center,
              font=\scriptsize,inner sep=3.2pt,minimum height=8.4mm,text width=25mm},
  hashbox/.style={box,fill=orange!10,draw=orange!55!black},
  lbl/.style={font=\scriptsize\itshape,inner sep=1pt},
  ar/.style={-{Stealth[length=1.6mm]},draw=black!55,line width=.5pt}
]
\node[box] (src) {\textbf{Frozen source}\\[1pt]novel text\\\texttt{1,073,378 chars}};
\node[hashbox, right=of src] (sha) {\textbf{SHA-256}\\[1pt]recorded in the\\manifest};
\node[box, right=of sha] (block) {\textbf{Block}\\[1pt]1,500 chars\\100 overlap};
\node[box, right=of block] (span) {\textbf{Verified span}\\[1pt]exact characters\\+ \texttt{start}/\texttt{end}};
\node[box, right=of span] (rel) {\textbf{Relation}\\[1pt]typed edge\\+ evidence string};
\node[box, right=of rel] (ans) {\textbf{Answer}\\[1pt]chosen option\\+ cited passage ids};

\draw[ar] (src) -- (sha);
\draw[ar] (sha) -- (block);
\draw[ar] (block) -- (span);
\draw[ar] (span) -- (rel);
\draw[ar] (rel) -- (ans);
\node[lbl, below=1.6mm of span] {offsets must slice back to the same string};
\node[lbl, below=1.6mm of ans] {every citation resolves or is reported};

\begin{scope}[on background layer]
\node[draw=blue!25,dashed,rounded corners=3pt,fit=(src)(sha)(block)(span),inner sep=5pt,
      label={[font=\scriptsize\bfseries,blue!45!black]above:build-time audit}] {};
\node[draw=orange!45!black,dashed,rounded corners=3pt,fit=(rel)(ans),inner sep=5pt,
      label={[font=\scriptsize\bfseries,orange!45!black]above:scoring-time audit}] {};
\end{scope}
\end{tikzpicture}}
\caption{Provenance chain for one graph edge. A relation is admitted only if its evidence
string occurs verbatim in a specific passage, which is itself an offset-bounded slice of
the frozen source file whose SHA-256 is in the manifest. The audit walks the chain in
both directions: from a scored answer back to the characters it quotes, and from any
source character to the nodes and edges that claim it.}
\label{fig:provenance}
\end{figure*}
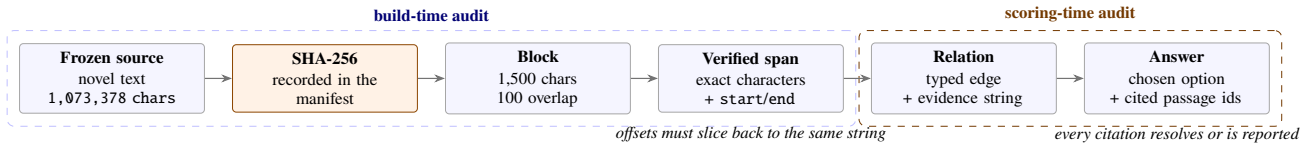

\section{Graph Construction and Use}
\label{sec:build}

\subsection{Building the index}
\label{sec:pipeline}

\begin{figure}[t]
\centering
\includegraphics[width=\columnwidth]{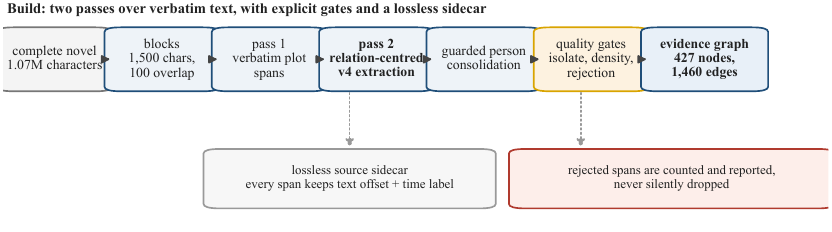}
\caption{The build. Two passes over verbatim text separate plot-bearing spans from
literary padding, then extract typed relations from the survivors. Gates are explicit,
rejections are counted, and the complete novel is retained as a lossless sidecar.}
\label{fig:pipeline}
\end{figure}

The graphs under test were built by two pipelines, but both follow the same two-pass
logic, shown in Figure~\ref{fig:pipeline}. The complete novel is split into blocks of
1,500 characters with 100 characters of overlap. Pass one asks a model to select the
spans that carry plot information and to leave literary description, filler dialogue and
scene setting behind; every selected span is copied from its block and checked back
against it, and a block whose selection fails validation is kept whole and flagged
rather than quietly dropped. Pass two extracts typed entities and relations from the
verified spans, with each relation carrying the evidence string that justifies it.
Person-name variants are then consolidated only when a supplied quote supports the
merge, and the result passes three explicit gates. The isolated-node rate must stay at or below 60\%,
edges per node at least 0.5, and rejected-relation rate at most 55\%.

Nothing is thrown away. The complete novel stays beside the graph as a lossless sidecar,
and each retained span keeps its character offsets in that source. A rejected span is
counted in the run metadata, which is how we can report elsewhere that 164 quotes and 91
relations were refused during the public rebuild of one novel while the final graph
still passed every gate.

\subsection{Four ways to use a graph}
\label{sec:routes}

\begin{figure*}[t]
\centering
\includegraphics[width=\textwidth]{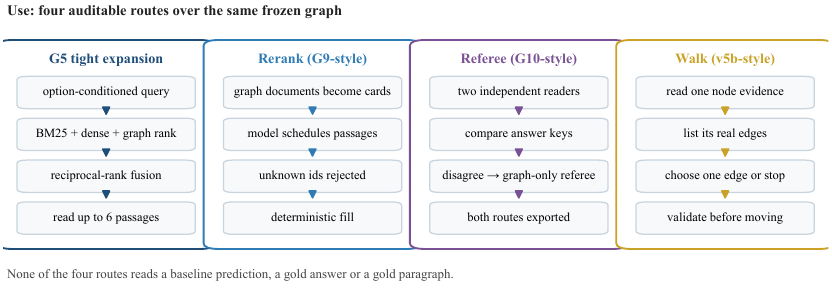}
\caption{The four route families behind the five graph conditions. Fusion and reranking
select a dossier before the reader sees it; refereeing compares two independent readers
and arbitrates only on disagreement; walking hands the choice of the next edge to the
reader and validates every move against the adjacency list. No route has access to a
baseline prediction, a gold answer or a gold paragraph.}
\label{fig:methods}
\end{figure*}

We evaluate five graph conditions. Three of them (G1--G3) apply deterministic
option-order permutations to the same evidence, which isolates the effect of ordering
from the effect of structure; G4 is their graph-only majority vote, and G5 is a tighter
expansion route that produces the strongest pooled accuracy in this study.
Figure~\ref{fig:methods} sketches the four route families that the conditions instantiate.

\begin{itemize}[leftmargin=1.05em,itemsep=1.5pt,topsep=2pt]
\item \textbf{Fusion (G5).} Each option supplies its own query. Keyword search, dense
retrieval and a graph ranking fused by reciprocal rank produce a short dossier; the
reader sees at most six passages.
\item \textbf{Reranking.} Graph documents become candidate cards, and the model
schedules which passages to read. Identifiers that do not exist in the graph are
rejected, and empty slots are filled deterministically.
\item \textbf{Refereeing.} Two readers answer independently. When their answer keys
disagree, a referee sees the union of their evidence and both proposals, and never sees
a baseline prediction or a gold answer.
\item \textbf{Walking.} The reader observes one node's evidence, is shown that node's
real adjacent edges, chooses one edge or stops, and repeats. Every move is validated
against the adjacency list before it is taken.
\end{itemize}

No graph condition has access to a baseline prediction, a gold answer or a gold
paragraph. Archived records for G5 carry an explicit \texttt{baseline\_access=false}
flag. One caveat belongs here rather than in a footnote. G4 and G5 were selected while
we were inspecting this corpus, so all five graph conditions are exploratory. A separate
confirmatory protocol with 340 questions, a compute-matched flat baseline and a
pre-registered primary hypothesis exists and has not been opened; Section~\ref{sec:protocol}
describes it without results.

\subsection{Gold evidence stays out of the loop}
\label{sec:goldisolation}

Clue and answer paragraphs are loaded only after graphs are frozen and predictions are
written. They are used to score evidence overlap and to compute the audit statistics in
Section~\ref{sec:evidence}, never to retrieve, rerank, route or calibrate an answer. A
node counts as gold-overlap when its stored evidence shares a normalized substring of at
least eight characters with an annotated paragraph. The rule is mechanical, and the
complete node-level table is released so that a reader can disagree with individual
calls.

\section{Evaluation Protocol}
\label{sec:protocol}

\begin{figure*}[t]
\centering
\includegraphics[width=\textwidth]{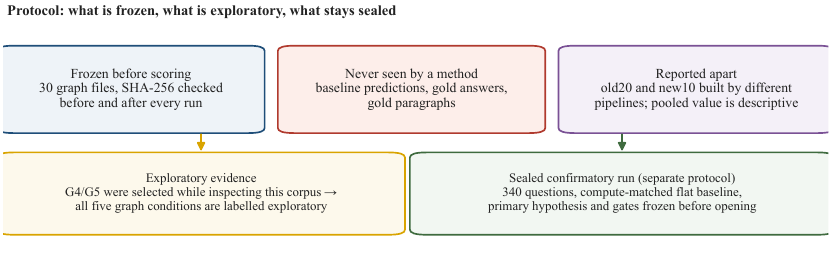}
\caption{What the protocol freezes, what no method may see, what must be reported apart,
and what remains sealed. The pooled thirty-novel value is descriptive because the two
cohorts come from different build pipelines; the confirmatory run follows a separate
protocol whose answers have not been opened.}
\label{fig:protocol}
\end{figure*}

Figure~\ref{fig:protocol} states the protocol in one picture: what is frozen, what no
method may see, what is reported apart, and what remains sealed.

\paragraph{Baselines.}
B1 reads the final 50,000 characters of the novel, the natural linear history when a
context window runs out. B2 compresses the whole book hierarchically inside the same
answer budget. B3 runs ordinary vector retrieval over contiguous chunks with no graph
links. \qzero\ supplies only the question and its options, which measures how much of
the set can be answered from general knowledge or from the question wording alone.
\qhard\ is the subset that \qzero\ answers incorrectly; it removes items a model can
answer without the book, and it changes nothing about what any method receives.

\paragraph{Statistics.}
Accuracy is reported as a micro average over questions and as an equal-weight
novel-macro average. Uncertainty uses a 95\% Wilson interval per condition and a
5,000-resample bootstrap clustered by novel for paired differences. Graph-versus-baseline
comparisons use exact McNemar tests on the same questions, with Holm correction over the
fifteen planned contrasts. Every question stays in the denominator; no item is dropped
for being hard.

\paragraph{The sealed run.}
The confirmatory protocol fixes a primary hypothesis before any scoring. On 340
questions, one graph route must exceed a compute-matched flat retrieval baseline by at
least five percentage points, with a novel-cluster bootstrap interval that excludes
zero. Answers stay sealed until the protocol's governance gates are satisfied, and
nothing from that run appears in this paper. We mention it because the difference
between the two studies is the point. This one is exploratory and says so.

\section{Results}
\label{sec:results}

\subsection{Main accuracy}
\label{sec:mainacc}

\begin{figure*}[t]
\centering
\includegraphics[width=\textwidth]{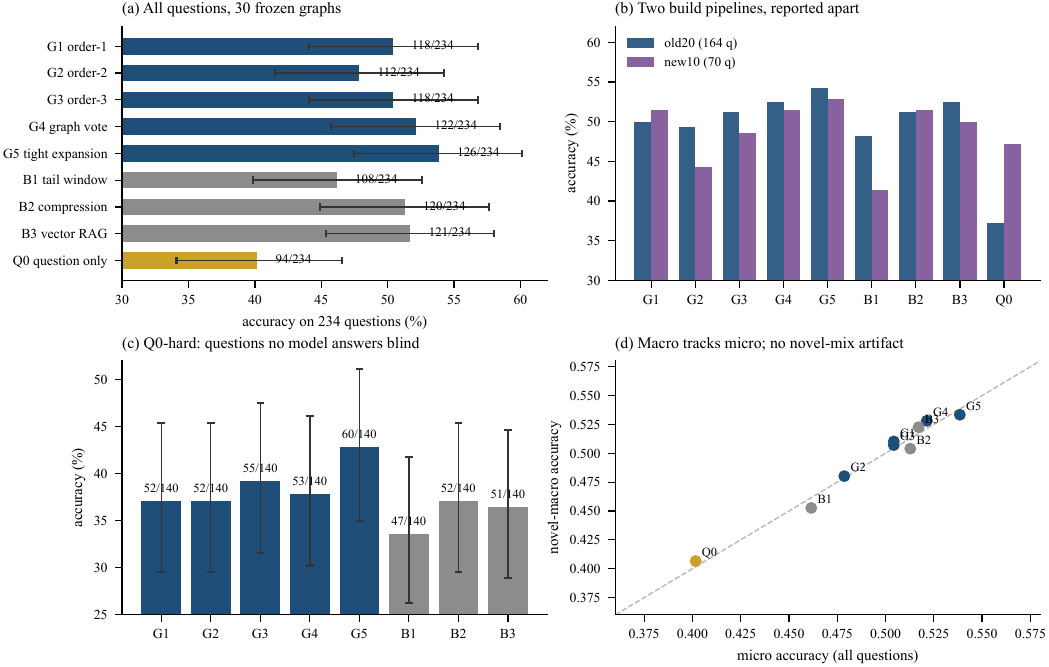}
\caption{Accuracy by condition. (a) All 234 questions with Wilson intervals and correct/total
counts. (b) The same conditions split by build cohort. (c) The \qhard\ subset, where no
model can answer from the question alone. (d) Novel-macro accuracy against micro accuracy;
the diagonal is the identity. Panel (b) shows the largest single source of variance in the
study. It belongs to the graphs themselves.}
\label{fig:mainacc}
\end{figure*}

Table~\ref{tab:main} gives every condition on all 234 questions, and
Figure~\ref{fig:mainacc} shows the same numbers with intervals, split by cohort, restricted
to the \qhard\ subset, and compared as micro versus novel-macro averages. The strongest
graph route, G5, answers 126 questions correctly (53.85\%, 95\% Wilson interval
[47.4, 60.1]). The recent-window baseline answers 108 (46.15\%), compression B2
answers 120 (51.28\%), vector retrieval B3 answers 121 (51.71\%), and the question-only
control \qzero\ answers 94 (40.17\%). The graph-only majority vote G4 sits between them
at 122 (52.14\%).

The gap between the graph routes and the question-only control is the one comparison
that needs no statistics to interpret. A model that never sees the novel answers 40\% of
this benchmark; the same model with graph-selected evidence answers 54\%. That is the
same reading of the corpus for both, and it is where the study's practical value sits.

Novel-macro averages track the micro values closely for every condition (Table~\ref{tab:main},
fourth column), so no condition wins by being lucky on the novels that carry more questions. The differences between the two cohorts
in Figure~\ref{fig:mainacc}(b) are larger than the differences between methods within a
cohort, and Section~\ref{sec:evidence} accounts for that spread.

\begin{table*}[t]
\centering
\caption{Accuracy on the frozen thirty-novel evaluation (234 questions, one local
\texttt{qwen3.5:9b} reader, reasoning disabled). Pooled values are descriptive because the
two cohorts were produced by different build pipelines. Wilson intervals are per
condition; the \qhard\ column restricts to the 140 questions the question-only control
answers incorrectly. Every value is regenerated from the archived per-question table.}
\label{tab:main}
\begin{tabular}{lcccccc}
\toprule
Condition & Correct & Accuracy & 95\% Wilson & Novel macro & Q0-hard & vs.\ Q0 \\
\midrule
G1 order-1 & 118/234 & 50.43 & [44.1, 56.8] & 51.03 & 52/140 & +10.26 \\
G2 order-2 & 112/234 & 47.86 & [41.5, 54.2] & 48.02 & 52/140 & +7.69 \\
G3 order-3 & 118/234 & 50.43 & [44.1, 56.8] & 50.71 & 55/140 & +10.26 \\
G4 graph majority & 122/234 & 52.14 & [45.8, 58.5] & 52.81 & 53/140 & +11.97 \\
\textbf{G5 tight expansion} & 126/234 & \textbf{53.85} & [47.4, 60.1] & \textbf{53.34} & 60/140 & +13.68 \\
B1 tail window & 108/234 & 46.15 & [39.9, 52.6] & 45.27 & 47/140 & +5.98 \\
B2 compression & 120/234 & 51.28 & [44.9, 57.6] & 50.40 & 52/140 & +11.11 \\
B3 vector RAG & 121/234 & 51.71 & [45.3, 58.0] & 52.26 & 51/140 & +11.54 \\
Q0 question only & 94/234 & 40.17 & [34.1, 46.6] & 40.66 & -- & +0.00 \\
\bottomrule
\end{tabular}

\end{table*}

\subsection{Per-novel variance dominates the method ranking}
\label{sec:heatmap}

\begin{figure*}[t]
\centering
\includegraphics[width=\textwidth]{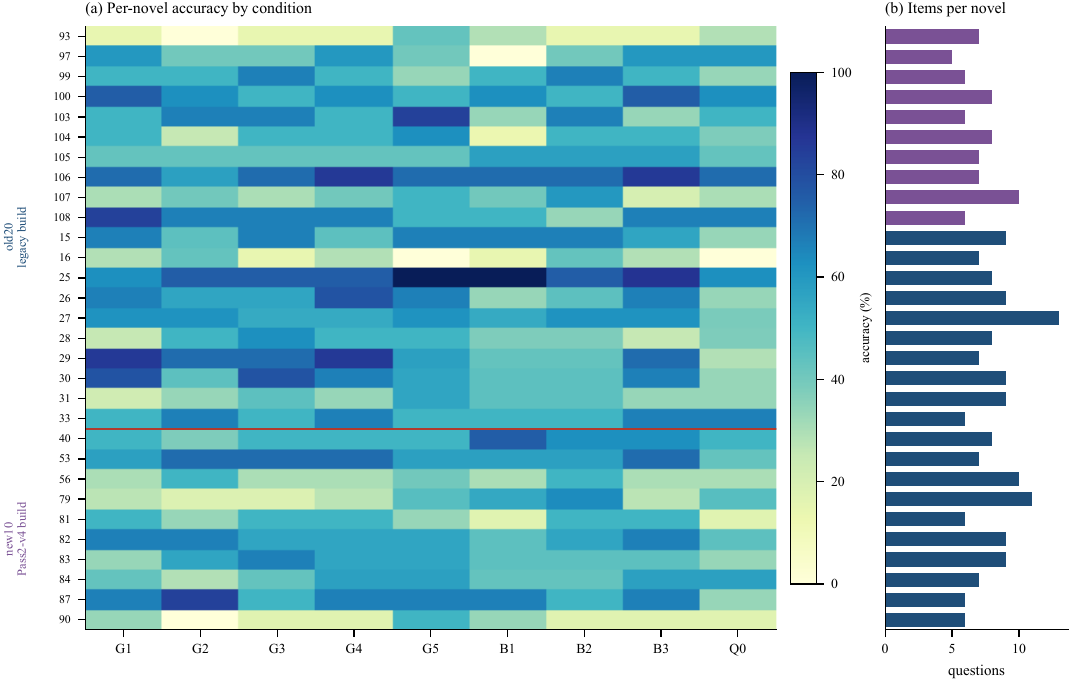}
\caption{Per-novel accuracy for every condition. (a) Rows are novels in cohort order, with
the cohort boundary marked; columns are conditions. (b) Questions per novel. Cells are
computed from the released per-question table, so the heatmap and the pooled table cannot
disagree. The old20 block is visibly weaker for the graph routes, and its items are also
unevenly distributed across novels.}
\label{fig:heatmap}
\end{figure*}

Figure~\ref{fig:heatmap} breaks accuracy down by novel. G5 accuracy ranges from 0\% to 100\%
across the thirty books, with a median of 53\% and an interquartile range of 45--62\%.
Several novels are answered correctly by every condition, and several by none. Method
differences of two to four percentage points are small relative to that spread, which is
why a single pooled number should not be read as a stable ranking.

\subsection{Paired comparisons, and what survives correction}
\label{sec:paired}

\begin{figure}[t]
\centering
\includegraphics[width=\columnwidth]{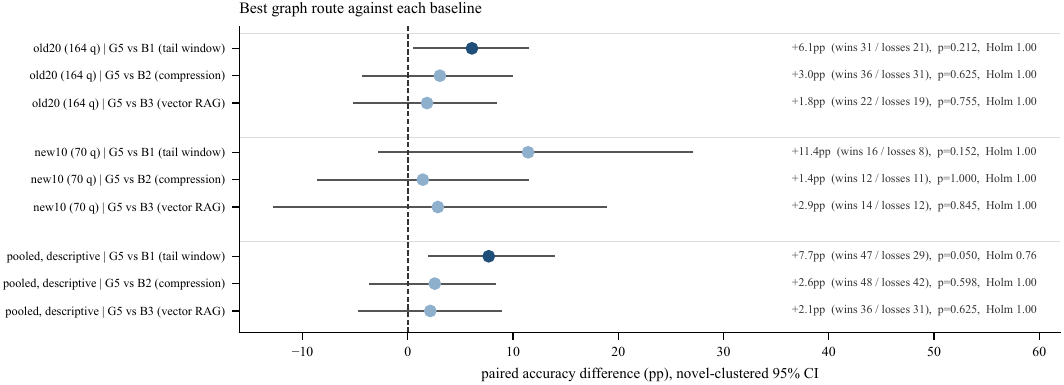}
\caption{Paired differences between the strongest graph route and each baseline. Points are
paired accuracy differences in percentage points; bars are novel-clustered 95\% bootstrap
intervals; annotations give wins, losses, exact McNemar $p$ and the Holm-adjusted value
across the fifteen planned graph--baseline contrasts. Nothing survives correction, so the
contrast with the recent window points at where to look next.}
\label{fig:pairwise}
\end{figure}

Figure~\ref{fig:pairwise} reports the paired contrast between G5 and each baseline on the
same questions, with novel-clustered bootstrap intervals and exact McNemar tests. Against
the recent window the gap is $+7.69\pp$ pooled (47 wins, 29 losses, $p=0.0505$), and it is
larger inside the newer cohort ($+11.43\pp$, $p=0.1516$) than inside the older one
($+6.10\pp$, $p=0.2116$). The contrasts against compression ($+2.56\pp$) and vector
retrieval ($+2.14\pp$) are smaller and their intervals cross zero. Holm correction over the
fifteen planned contrasts leaves no surviving comparison; the adjusted value for the
tail-window contrast is $0.76$.

\subsection{Where the graph is the only winner}
\label{sec:census}

\begin{figure*}[t]
\centering
\includegraphics[width=\textwidth]{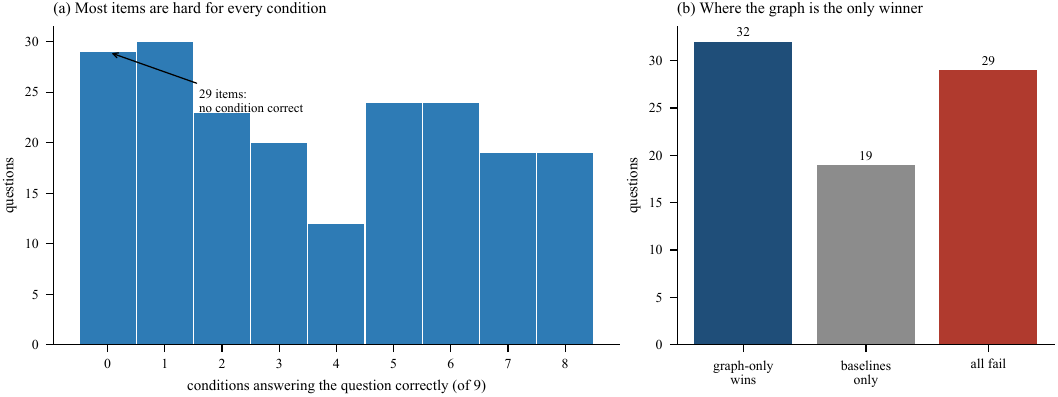}
\caption{Difficulty census across the nine conditions. (a) How many conditions answer each
question correctly. The spike at zero counts real failures.
(b) Questions where a graph condition is the only correct answer, where a baseline is the
only correct answer, and where nothing is correct.}
\label{fig:difficulty}
\end{figure*}

Counting per question rather than per average, 32 of 234 items are answered correctly by at
least one graph condition while all three non-graph baselines miss them. Another 29 items
are answered by nobody. Figure~\ref{fig:difficulty} shows the distribution of how many
of the nine conditions solve each question. Most items are hard for every condition, which
is the honest summary of a detective benchmark. The interesting cases are not the ones
where one method beats another by a few points, but the smaller set where structure changes
the outcome.

\subsection{Evidence audit: index quality limits answer quality}
\label{sec:evidence}

\begin{figure*}[t]
\centering
\includegraphics[width=\textwidth]{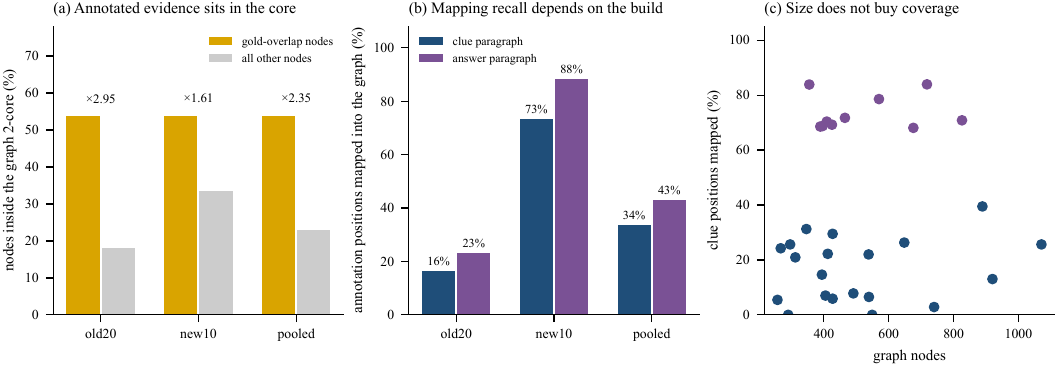}
\caption{Evidence audit. (a) Share of nodes inside the graph 2-core, split by whether the
node overlaps an annotated paragraph; multipliers give the enrichment ratio. (b) Share of
annotated clue and answer positions that a graph represents at all, by cohort. (c) Graph
size against annotation mapping rate. The larger graphs are not the better-mapped ones.
Panel (a) is an association between annotation overlap and topology; it is not evidence
about model attention or about answer accuracy.}
\label{fig:goldaudit}
\end{figure*}

Figure~\ref{fig:goldaudit} separates two questions that pooled accuracy hides. First, does
annotated evidence sit where the graph is structurally dense? Yes, and consistently:
53.66\% of gold-overlap nodes fall inside the 2-core for old20 versus 18.18\% of other
nodes ($2.95\times$ enrichment, Haldane odds ratio 5.20), and 53.87\% versus 33.44\% for
new10 ($1.61\times$, odds ratio 2.32). Pooled enrichment is $2.35\times$.

Second, does the pipeline map the annotation into the graph at all? Here the cohorts
diverge sharply. old20 maps 16.3\% of clue positions and 23.2\% of answer positions into
its graphs; new10 maps 73.2\% and 88.2\%. Pooled, those are 33.7\% and 43.0\%. A graph
can be topologically well formed and still be missing three quarters of the evidence a
question needs, and an accuracy comparison that pools the two builders averages a
pipeline defect into a method score.

\subsection{Graph inventory}
\label{sec:inventory}

\begin{figure*}[t]
\centering
\includegraphics[width=\textwidth]{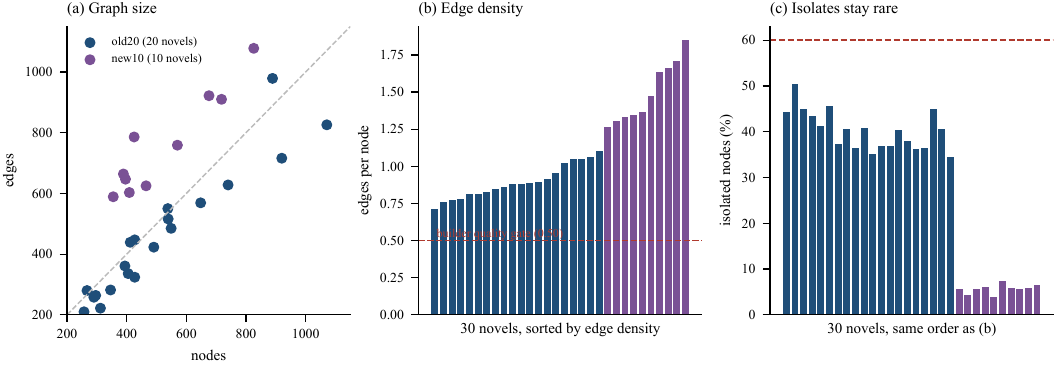}
\caption{The frozen graphs as inputs. (a) Nodes against edges with the identity line; points
below the diagonal are the sparse legacy graphs. (b) Edge density per novel, with the 0.50
builder gate marked; novels are sorted by density and colours give the cohort. (c) Isolated
nodes in the same order. Two cohorts built by different pipelines differ in density by a
factor of four, which is a property of the builders and a candidate explanation for the
cohort gap in Figure~\ref{fig:goldaudit}(b).}
\label{fig:inventory}
\end{figure*}

Figure~\ref{fig:inventory} describes the thirty frozen graphs as inputs rather than as
results. Edge density ranges from 0.49 to 2.05 edges per node, and seven novels fall below
the 0.5 gate that the later builder enforces, all of them from the legacy cohort. Isolated
nodes are rare in both cohorts (median 4.7\%, maximum 51\%), so the sparse graphs are sparse
in their connections rather than broken in structure.

\subsection{An oracle ceiling that moves with the option letter}
\label{sec:oracle}

\begin{figure*}[t]
\centering
\includegraphics[width=\textwidth]{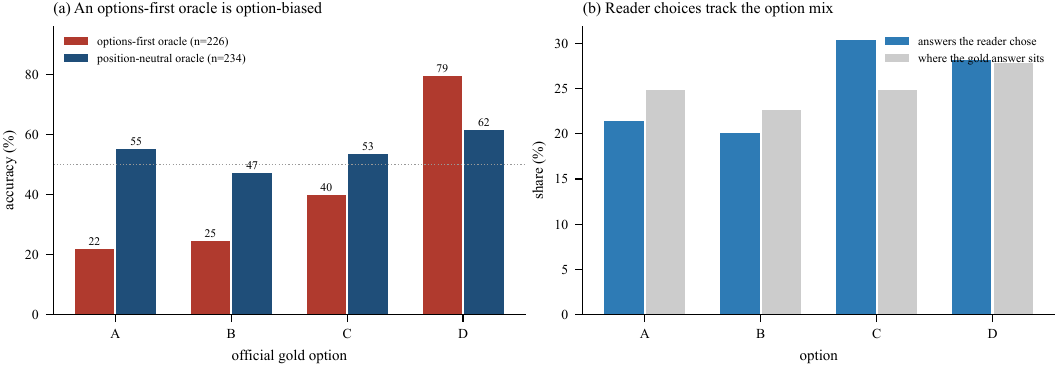}
\caption{Option-position effects. (a) Oracle accuracy by gold option letter for an
options-first oracle and a position-neutral one. (b) Which options the graph reader chose,
against where the gold answers actually sit. Any single oracle number is therefore a weak
ceiling: it can be inflated by the option layout, and the position-neutral variant is the
one worth quoting.}
\label{fig:q0artifact}
\end{figure*}

One more measurement belongs in the results because it disciplines how the others are read.
An options-first oracle, which shows the model every official clue paragraph together with
the options in their original order, is correct on 79\% of questions whose gold answer is
option D and on 22--53\% of questions whose gold answer is A, B or C
(Figure~\ref{fig:q0artifact}a). A position-neutral oracle, which prompts for a letter
without the option list, flattens that curve to 47--62\%. The reader's own choices follow
the gold option mix rather than a fixed letter
(Figure~\ref{fig:q0artifact}b). The artifact belongs to the oracle.

\section{Discussion}
\label{sec:discussion}

\paragraph{What the numbers support.}
Reorganizing a novel into a walkable graph gives a small local model a measurable edge
over reading the last fifty thousand characters. It reaches 53.85\% against 46.15\%, and 42.86\%
against 33.57\% on the questions that no model answers from the question alone. On this
benchmark the graph routes also beat whole-book compression and ordinary vector retrieval,
but by two to three percentage points, which is inside the noise of a 234-question set and
inside the between-novel spread documented in Figure~\ref{fig:heatmap}. The one comparison
that comes close to conventional significance, the recent-window contrast, does not survive
correction across the planned family. We therefore report the design as promising and the
effect as unproven.

\paragraph{The index is the first bottleneck, and it is measurable.}
The evidence audit gives the more useful result. Annotated evidence does concentrate in the
topological core of these graphs, but the pipeline that turns annotation into graph nodes
recovers anywhere from 16\% to 73\% of clue positions depending on which builder produced the
graph. Nothing about the retrieval routes can fix a graph that never contained the evidence.
This is why the cohort split matters more than the method ranking. Pooling the cohorts would
have averaged a build defect into a retrieval score and left the reader to guess which was
which.

\paragraph{Where the graph earns its keep.}
Thirty-two questions are answered by a graph condition while every non-graph baseline misses
them, and twenty-nine are answered by nobody. The second number is a reminder that a
detective benchmark contains items that are hard for reasons no retrieval design addresses:
ambiguous annotation, translation artifacts, or reasoning that a 9B reader cannot complete
even with the right paragraphs in front of it. The first number is the part worth chasing.
If the graph-only wins turn out to cluster in questions that require chaining evidence across
distant passages, that would be a cleaner claim than any pooled average, and it is the
analysis we would run next.

\paragraph{The option layout moves an oracle ceiling by fifty points.}
An options-first oracle scores 79\% when the gold answer is option D and 22--53\% when it is A,
B or C. Any ceiling quoted from such a run measures the layout.
The position-neutral variant of the same oracle flattens to 47--62\%, and it is the number
worth trusting. We mention this at length because an inflated ceiling makes a graph method
look closer to solved than it is, and the correction changes the interpretation of every
comparison against it.

\paragraph{Practical reading.}
For someone building this kind of index, freeze one builder, report annotation mapping recall
alongside accuracy, and evaluate on novels the builder never saw. Our two cohorts differ in
mapping recall by a factor of four, and any of the three effects we measured is smaller than
that factor.

\begin{table*}[t]
\centering
\caption{Accuracy by cohort and the preservation of questions the question-only control
already answers. The pooled column is descriptive, because the two cohorts differ in both builder
and mapping recall.}
\label{tab:cohort}
\small
\begin{tabular}{lcccc}
\toprule
Condition & old20 (164) & new10 (70) & pooled (234) & preserved on Q0-correct \\
\midrule
G1 order-1 & 50.0 & 51.4 & 50.4 & 70.2 \\
G2 order-2 & 49.4 & 44.3 & 47.9 & 63.8 \\
G3 order-3 & 51.2 & 48.6 & 50.4 & 67.0 \\
G4 graph majority & 52.4 & 51.4 & 52.1 & 73.4 \\
G5 tight expansion & 54.3 & 52.9 & 53.8 & 70.2 \\
B1 tail window & 48.2 & 41.4 & 46.2 & 64.9 \\
B2 compression & 51.2 & 51.4 & 51.3 & 72.3 \\
B3 vector RAG & 52.4 & 50.0 & 51.7 & 74.5 \\
Q0 question only & 37.2 & 47.1 & 40.2 & -- \\
\bottomrule
\end{tabular}

\end{table*}

\section{Limitations}
\label{sec:limits}

The sample is thirty novels and 234 questions, which is enough to see large effects and not
enough to resolve two-point differences; the Wilson intervals in Table~\ref{tab:main}
overlap across most conditions. The two cohorts come from different pipelines, so no pooled
number identifies a method effect on its own, and we report the split rather than adjust for
it. G4 and G5 were selected while inspecting this corpus, and if we had chosen differently
the pooled ranking might look different; the sealed confirmatory protocol exists to settle
exactly that, and it has no results yet.

Gold-overlap uses lexical matching with an eight-character floor, which misses paraphrases
and counts some incidental overlaps; the released node-level table lets a reader re-run the
comparison with a different rule. Some legacy runs preserve predictions and signatures, and lack
exact tokenizer counts or per-call timings, and those fields are marked unavailable
rather than reconstructed. One local 9B reader and one English-translated Chinese corpus
bound the external validity; we make no claim about larger models or other genres. Finally,
the force-directed view we use for continuity with an earlier presentation is
seed-dependent, and its geometry is not evidence about narrative time or semantic distance.

\section{Reproducibility}
\label{sec:repro}

Every number in this paper is regenerated from committed machine outputs rather than typed
by hand. The per-condition summaries and paired tests come from \path{generated/dqa30_frozen_results.json} for the per-condition summaries and
paired tests, \path{generated/dqa30_per_question.csv} for the 234$\times$9 answer matrix,
\path{generated/dqa30_fair_gold_results.json} for the oracle variants,
\path{generated/dqa30_gold_dense_regions.json} for the 2-core and mapping statistics, and
\path{config/dqa30_frozen_graphs.json} for the graph inventory with SHA-256 hashes. The
figures in this paper are rebuilt by \path{make_figures.py} and the tables by
\path{tables/make_tables.py}, both of which read those files directly, so a change in the
data propagates to the figures and tables on the next build. The manuscript compiles with
\path{tectonic -X compile main.tex}; it needs no local TeX installation beyond that single
binary.

The per-question table is the audit surface. Each row gives the cohort, novel, question id,
official gold option and the option each condition chose, which is enough to recompute every
accuracy, every interval and every paired test in this paper, and to check our counts of the
questions where only a graph route succeeds.

\section{Conclusion}
\label{sec:conclusion}

A graph turns a novel into something a small model can walk instead of merely read, and on
this benchmark that is worth about eight percentage points against the linear history a
context window leaves behind. The effect does not survive correction, so we do not present it
as settled. What does hold is the audit. Annotated evidence is measurably concentrated in
these graphs' cores, the pipelines that build them differ by a factor of four in how much of
that annotation they capture, and an oracle ceiling can move fifty points with the option
layout. Those three measurements are the reason this evaluation separates index quality from
answer accuracy, and they are the parts we would carry into the next study.

\balance
\AtBeginEnvironment{thebibliography}{\raggedright}

\end{document}